\documentclass[letterpaper, 10 pt, conference]{ieeeconf}  

\IEEEoverridecommandlockouts                              

\title{\LARGE \bf
DispSegNet: Leveraging Semantics for End-to-End Learning of Disparity Estimation from Stereo Imagery
}

\author{Junming Zhang$^{1}$, Katherine A. Skinner$^{2}$, Ram Vasudevan$^{3}$ and Matthew Johnson-Roberson$^{4}$%
\thanks{
$^{1}$J. Zhang is with the Department of Electrical Engineering and Computer Science, University of Michigan, Ann Arbor, MI 48109 USA {\tt\small junming@umich.edu}
\newline \indent $^{2}$K. Skinner is with the Robotics Program, University of Michigan, Ann Arbor, MI 48109 USA {\tt\small kskin@umich.edu}
\newline \indent $^{3}$R. Vasudevan is with the Department of Mechanical Engineering, University of Michigan, Ann Arbor, MI 48109 USA {\tt\small ramv@umich.edu}
\newline \indent $^{4}$M. Johnson-Roberson is with the Department of Naval Architecture and Marine Engineering, University of Michigan, Ann Arbor, MI 48109 USA {\tt\small mattjr@umich.edu}}
}

\usepackage{perl_acronyms}
\usepackage{graphicx}
\usepackage{amsmath}
\usepackage{tabularx}
\usepackage{float}
\usepackage{xcolor}
\usepackage{graphicx}
\usepackage{flushend}
\usepackage{textcase}
\usepackage[tablename=TABLE]{caption}
\DeclareCaptionTextFormat{up}{\MakeTextUppercase{#1}}
\usepackage[caption=false]{subfig}
\usepackage[
    style=ieee,
    doi=false,
    isbn=false,
    url=false,
    eprint=false,
    backend=bibtex,
    natbib=true
    ]{biblatex}
    
\newcommand{\matt}[1]{}
\renewcommand{\matt}[1]{{\color{red} MattJR: {#1}}}
\newcommand{\katie}[1]{}
\renewcommand{\katie}[1]{{\color{red} Katie: {#1}}}
\begin{document}

\maketitle
\begin{abstract}
Recent work has shown that convolutional neural networks (CNNs) can be applied successfully in disparity estimation, but these methods still suffer from errors in regions of low-texture, occlusions and reflections. Concurrently, deep learning for semantic segmentation has shown great progress in recent years. In this paper, we design a CNN architecture that combines these two tasks to improve the quality and accuracy of disparity estimation with the help of semantic segmentation. Specifically, we propose a network structure in which these two tasks are highly coupled. One key novelty of this approach is the two-stage refinement process. Initial disparity estimates are refined with an embedding learned from the semantic segmentation branch of the network. The proposed model is trained using an unsupervised approach, in which images from one half of the stereo pair are warped and compared against images from the other camera. Another key advantage of the proposed approach is that a single network is capable of outputting disparity estimates and semantic labels. These outputs are of great use in autonomous vehicle operation; with real-time constraints being key, such performance improvements increase the viability of driving applications. Experiments on KITTI and Cityscapes datasets show that our model can achieve state-of-the-art results and that leveraging embedding learned from semantic segmentation improves the performance of disparity estimation.
\end{abstract}

\section{INTRODUCTION}
Disparity estimation is an important problem in low-level vision. Given two stereo rectified images, disparity refers to the relative horizontal displacement of two corresponding pixels in the left and right images. 
From dense disparity maps, we can estimate three dimensional geometry, which is critical for many computer vision applications, including autonomous vehicle navigation and 3D model reconstruction.

Traditionally, dense disparity has been estimated using window-based correlation, with smoothing, occlusion and globally-optimal matching constraints applied~\cite{jang2011efficient,adhyapak2007stereo,kanade1991stereo,hirschmuller2008stereo}. However, it is difficult to hand-craft these constraints. Additionally, global optimization is not practical for real-time applications. Recently, with the help of \acp{CNN}, stereo matching has greatly advanced. Meaningful features learned by \acp{CNN} prove to be more effective than hand-crafted ones. More sophisticated architectures are able to estimate dense disparity through end-to-end training. This end-to-end disparity regression from stereo pairs requires a large amount of image pairs with ground truth disparities during training. However, there is currently no such real dataset available. Instead, models are pretrained on large synthetic datasets~\cite{mayer2016large,dosovitskiy2015flownet} and then fine-tuned on the real-world target dataset. With this training pipeline, recent papers~\cite{kendall2017end,yang2018segstereo,chang2018pyramid,liang2018learning} achieve an impressive error rate below 2\% in the KITTI benchmark stereo matching task~\cite{geiger2012we,Menze2015object}. Still, there are challenges to training on synthetic data and testing on real data. In this paper, we focus on developing an unsupervised method to do stereo matching for dense disparity estimation to help overcome these challenges.

\begin{figure}[t!]
\vspace{6pt}
\centering
\subfloat[Input stereo images]{%
  \includegraphics[clip,width=\linewidth]{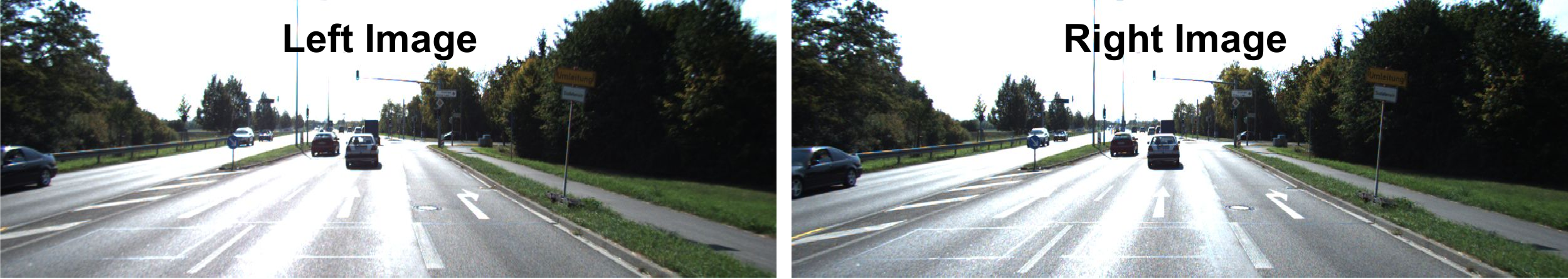}}%
  \vspace{-10pt}

\subfloat[Disparity prediction and error map without segment embedding]{%
  \includegraphics[clip,width=\linewidth]{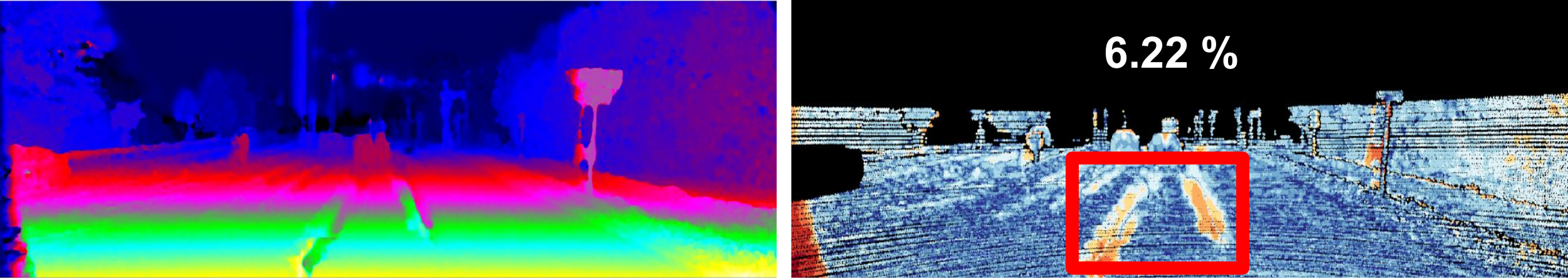}}%
  \vspace{-10pt}
  
\subfloat[Disparity prediction and error map with segment embedding]{%
  \includegraphics[clip,width=\linewidth]{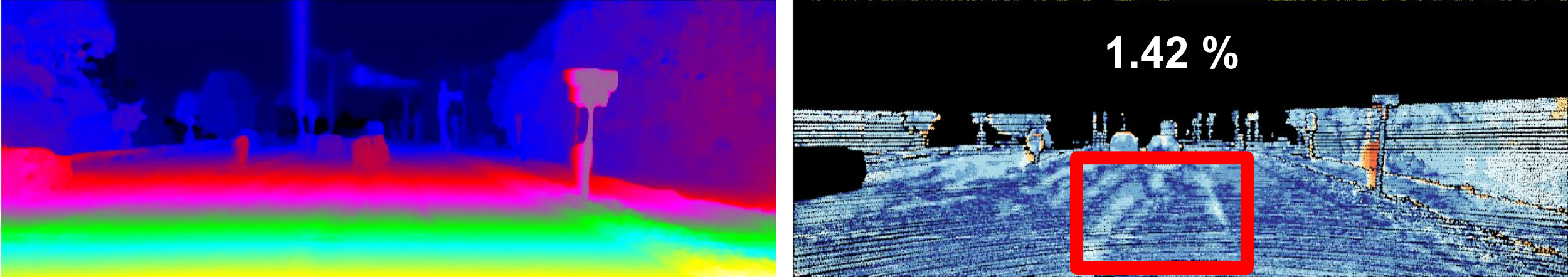}}%
  \vspace{-5pt}
\caption{Examples of advantage of fusing segment embedding into disparity estimation. With fused segment embedding, our model performs better in ill-posed regions, such as the area within the red bounding box in each figure. The white numbers in the error maps indicate percentage of incorrect pixels in all regions.}
\label{failure_case}
\vspace{-13pt}
\end{figure}

Despite advances in disparity estimation since the application of \acp{DNN}, finding correspondences in regions of high specularity, occlusions or low-texture regions is still a challenging problem. These areas manifest themselves as noise or missing regions in the resulting disparity map. For example, in Fig. \ref{failure_case}, the disparity for the center of the road is incorrect because it is an area of low-texture and it is hard to find correspondence in this region. We argue that more contextual semantic information is needed to determine accurate disparity in these challenging regions.

With the rise and success of object classification~\cite{krizhevsky2012imagenet,he2016deep}, a new task known as semantic segmentation has also gained popularity and benefited from access to large amounts of labeled data~\cite{lin2014microsoft,Cordts2016Cityscapes}. This problem moves beyond simple bounding boxes and attempts to assign every pixel in an image a semantic label. The dense nature of this problem is complimentary to the disparity estimation task. Moreover, segment embedding learned from semantic segmentation can provide further cues for estimating disparity within ill-posed regions, because disparity tends to be smooth within an object or segment. From this perspective, models for disparity estimation need to have a high-level understanding of objects or at least segments, so stereo matching is no longer a low-level vision problem. Here, we set out to exploit the connection between these two pixel labeling tasks -- disparity estimation and semantic segmentation -- to improve the performance for disparity estimation.


In this paper, we focus on unsupervised stereo matching guided with the semantic segmentation task. The main contributions of this paper are:

(1) We propose a model which outputs a disparity map and semantic segments simultaneously, and then both can be used to acquire 3D semantic information.

(2) We propose a structure and a smoothness loss which better fuses segment embeddings learned from the semantic segmentation task into the process of disparity estimation. Experiments show that these are helpful for disparity estimation.

(3) Our unsupervised model is able to achieve state-of-the-art results in the KITTI stereo vision benchmark dataset, and can also beat some supervised methods in certain regions. 

\section{RELATED WORK}

Typical stereo matching pipelines consist of four steps: matching cost computation, cost aggregation, disparity estimation and refinement. Traditional methods either use local descriptors to find the matching points within a predefined window~\cite{joglekar2014image}, or they minimize an energy function globally to get an optimal solution~\cite{hirschmuller2008stereo}. 

\textbf{Supervised Disparity Estimation.} Stereo matching has greatly advanced since \acp{CNN} were applied to this task by \citet{zbontar2016stereo}. That method was supervised, requiring large datasets with stereo images and disparity ground truth. With this supervised approach, after meaningful features are extracted from a deep Siamese architecture, a cost volume can be computed by simply concatenating features from both sides~\cite{zbontar2016stereo}, dot products~\cite{luo2016efficient,zbontar2016stereo}, a correlation function~\cite{mayer2016large}, or by concatenating all potential corresponding feature vectors from both sides~\cite{kendall2017end}. Several other papers have also focused on using information from cost volumes. They proposed different methods and structures, including simple convolutional layers~\cite{mayer2016large}, learning context from 3D convolution~\cite{kendall2017end}, using a spatial pyramid pooling module to incorporate more global context~\cite{chang2018pyramid}, a two-stage refinement structure~\cite{pang2017cascade} and two separate branches for small and large disparities~\cite{ilg2017flownet,liang2018learning}. In line with these suggestions, we form a five-dimensional cost volume by concatenating features from both sides and extracting information from it using 3D convolution. We then refine the initial disparity using extra information from segment embedding. 

Although some large datasets are now available for training in stereo matching, the size of available datasets is still relatively small compared to popular datasets for classification and detection. For example, KITTI 2012 and KITTI 2015, the most popular datasets for the stereo matching task, contain no more than 400 stereo images for training. In cases like this, unsupervised stereo matching has gained attention because it does not require ground truth disparity for training. Because of this, we focus on unsupervised learning in our approach for the stereo branch of our network. This maximizes the flexibility of the training sources, which is important because stereo ground truth is difficult to obtain.

\textbf{Unsupervised Disparity Estimation.} Deep unsupervised stereo matching relies heavily on warping error~\cite{garg2016unsupervised,godard2017unsupervised,zhou2017unsupervised,luo2018unsupervised}. This error is measured as the visual difference between a warped image from one half of a stereo pair and the real image from the other camera in the stereo setup. End-to-end training has become popular recently thanks to differentiable bilinear sampling, which can be used to warp images~\cite{godard2017unsupervised}. Additionally, a smoothness loss and left-right consistency loss also help improve the quality of results~\cite{godard2017unsupervised,zhong2017self}. Although results of these unsupervised methods are reasonable, a large performance gap still exists between these approaches and supervised methods. In this paper, we mainly focus on unsupervised stereo matching, and seek to use supervised semantic segmentation to help narrow this gap.

\textbf{Guided Disparity Estimation.} Both supervised and unsupervised stereo matching methods still have difficulty estimating correct disparity in flat, reflective and occluded regions. Thus, recent papers have sought to leverage extra information such as object-level knowledge~\cite{guney2015displets} and segment embedding~\cite{yang2018segstereo}. Their results show that exploiting available high-level information is useful for improving performance on the task of dense disparity estimation.

In this paper, we propose a fused model for semantic segmentation and disparity estimation that does not require ground truth disparity maps. Our proposed method is most similar to SegStereo~\cite{yang2018segstereo}, which was developed simultaneously  with our approach. However, our methods differ in several important ways. We focus on unsupervised stereo matching, where segment embedding is not only fused into disparity estimation, similar to SegStereo, but also is used to regularize disparity in the loss. Additionally, SegStereo computes a correlation layer, which may lose information, but we form a cost volume retaining all features, which enables the network to learn more complete feature representations. With additional refinement on the initial disparity, the results of our model outperform SegStereo by over a 2.5\% error rate on KITTI benchmark.

\begin{figure*}[t!]
  \vspace{5pt}
  \centering
  \includegraphics[width=0.9\linewidth]{{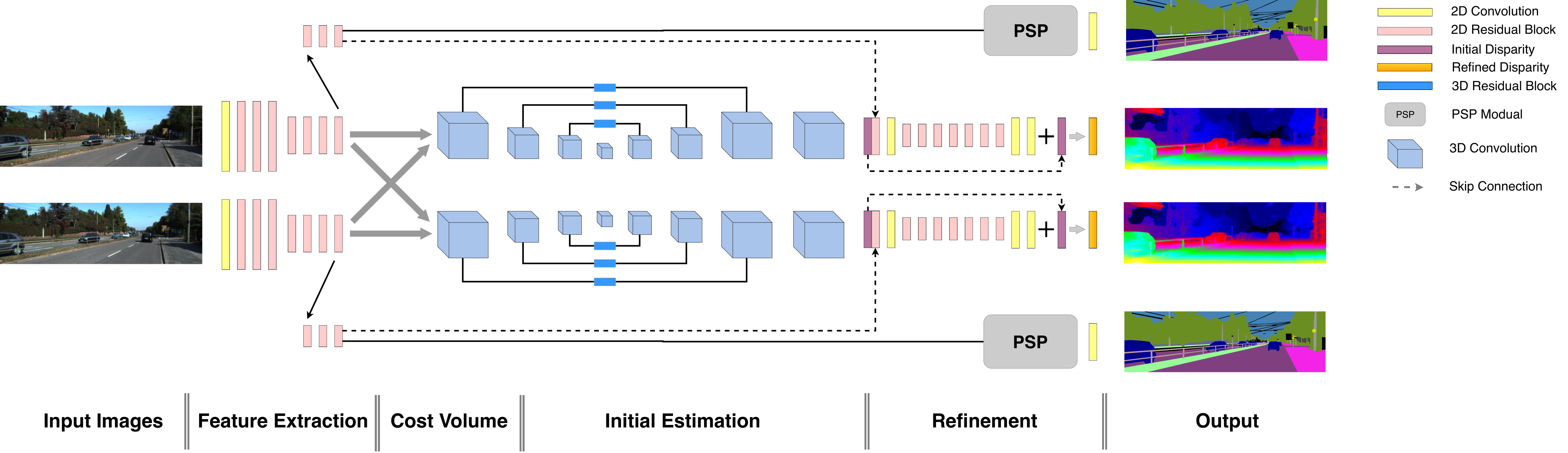}}
  \caption{Architecture of our model. The pipeline of our model consists of the following parts. (a) Input images: rectified input stereo images. (b) Feature extraction: useful features are extracted from input stereo images. (c) Cost volume: cost volume is formed by concatenating corresponding features from both sides. (d) Initial estimation: initial disparity is estimated from cost volume using 3D convolution. (e) Refinement: initial disparity is further improved by fusing segment embedding. The PSP module is used to incorporate more context information for the semantic segmentation task. (f) Output: estimated disparity and semantic segmentation from both left and right views are generated from the model. In this figure, 2D and 3D residual blocks are similar to identity blocks that are defined in the \cite{he2016deep}.}
  \label{fig:architecture}
  \vspace{-15pt}
\end{figure*}

\section{METHOD}
We present a joint model for disparity estimation and semantic segmentation. These two tasks are highly coupled in the network, with the semantic segment embedding being directly fused into the refinement process for disparity estimation. The whole architecture of our model is illustrated in Fig.~\ref{fig:architecture}. 

\subsection{Architecture for Disparity Estimation}
ResNet 50 structure [14] is used in the Siamese structure, which processes both the left and right images and generate high-level features for stereo matching and semantic segmentation. Features for the segmentation task come from deeper layers of the network than those used for the stereo matching task, as the former requires more contextual information than  the latter. Each task corresponds to a branch in the network. In the  disparity branch, the size of input features is 1/4 of original stereo images. We concatenate features for stereo matching from the left and  right viewpoints, and this produces a five-dimensional cost volume. An  eight-layer encoder-decoder with 3D convolution is then used to extract an initial disparity map from this volume. The structure of encoder-decoder is shown in Fig. 2 and the relative size of cube indicates the relative size of each layer and it outputs the initial disparity map. 3D transpose convolution is used in the decoder. Skip layers are processed by 3D residual blocks. The segment embedding is first resized to the same shape of original image and concatenated with the initial disparity map to do refinement. In details, the convolution layers in this paper refer to a convolution layer followed by a batch normalization layer and a leaky ReLU layer, except for the final output layer which only contain a regular convolution. The size of all kernels is 3 except for the first convolution layer in Siamese structure, which is 7. The 2D residual block is three layers deep and the 3D residual block is two layers deep.

\subsection{Cost Volume and Learning Context}
After calculating left and right features for stereo matching, we form a cost volume by concatenating them. Every feature vector from one side is concatenated with all potential corresponding feature vectors from the other side. This results in a cost volume with a dimensionality of Batchsize $\times$ (Max\_disparity+1) $\times$ Height $\times$ Width $\times$ Feature\_size. We form both left and right cost volumes to calculate a disparity for both views. Unlike other methods that use dot product or other metrics to measure correlation between feature vectors, the five-dimensional cost volume here enables the network itself to learn better correlation metric during training.


To extract information from the cost volume, a 3D convolution filter loops over all three dimensions of height, width and potential disparity values. This step captures broader contextual information. Since 3D convolution is memory intensive, an encoder-decoder structure is used to reduce the memory footprint. Soft argmin is used to produce the initial disparity map from this intermediate result.

\subsection{Disparity Refinement}
The initial disparity estimation contains too much noise and its accuracy is limited by error from poor matching in ill-posed regions, such as occluded, reflective and texture-less areas. However, the semantic segment embedding can be used to improve correspondence in those regions. Disparity in the ill-posed regions should have similar values as regions from the same semantic segment. Essentially, the same smoothness constraint that is often applied globally can more accurately be applied within object boundaries. To this end, after producing the initial disparity map, we use semantic segment information to refine the disparity. The residual structure of the refinement process is shown in Fig.~\ref{fig:architecture}.


After convergence, we assume the initial disparity is reasonable in most regions, so in the refinement stage we then focus on refining the disparity in ill-posed regions. The residual structure is used here and forces the model to learn this highly non-linear relationship in such regions. The initial disparity and the semantic segment embedding are concatenated as the input to later process. The output is then summed with the initial disparity to get the final estimation.

\subsection{Architecture for Semantic Segmentation}
In both the KITTI and Cityscapes~\cite{Cordts2016Cityscapes} datasets, only the left image from the stereo pair is labeled with ground truth semantic segments. However, we perform semantic segmentation on both images in the pair. Left disparity is used to warp the right semantic segmentation to the left view, which is in turn regularized by the left labels during training. Similar to PSPNet~\cite{zhao2017pyramid}, a PSP module is used to incorporate more contextual information from different scales. The size of input features to the PSP module is 1/8 of original stereo image. In the PSP module, input features are downsampled into three different sizes using averaging pooling, at scales of 1/2, 1/4 and 1/8 of the original input size. Then they are followed by a convolution with a 1x1 filter individually to reduce feature dimension to 1/4 of the original input feature dimension. Different scales of features are then concatenated after they are upsampled to the shape of the input feature space through bilinear interpolation. Finally it is followed by a 1x1 convolution to mix features at different scales.

\subsection{Loss Function}
For our approach, we pose stereo matching as an unsupervised problem. The object loss consists of three items that are defined as the following:

\begin{equation}
    Loss = \alpha_1 L_{init} + \alpha_2  L_{ref} + \alpha_3 L_{seg}
\end{equation}
\begin{equation}
    L_{init} = \beta_1 L_p + \beta_2 L_c + \beta_3 L_r
\end{equation}
\begin{equation}
    L_{ref} = \gamma_1 L_p + \gamma_2 L_c + \gamma_3 L_s
\end{equation}
where $L_{init}$ supervises initial estimated disparity, $L_{ref}$ supervises refined estimated disparity and $L_{seg}$ supervises predicted semantic segments. We set $\alpha_1=0.3$, $\alpha_2=0.7$, $\alpha_3=0.1$, $\beta_1=0.8$, $\beta_2=0.01$, $\beta_3=0.001$, $\gamma_1=0.8$, $\gamma_2=0.05$ and $\gamma_3=0.005$ during the training. The other terms are defined as follows:

\begin{itemize}
\item 
Photometric loss ($L_p$): Let $I_L$ and $I_R$ be the input left and right images, and $D_L$ and $D_R$ be the predicted left and right disparity maps. The warping function $F(I, D)$ is able to warp image $I$ to the other view based on the disparity map $D$ with bilinear sampling. The reconstructed left image is $I_{L}^{'} = F(I_R, D_L)$, and the reconstructed right image is $I_{R}^{'} = F(I_L, D_R)$. The reconstructed image should be very similar to the original input image. We use both Euclidean distance and a structure similarity term SSIM $S(\cdot)$ to improve the robustness in ill-posed regions~\cite{godard2017unsupervised}. For the left image, photometric loss is defined as follows:
\begin{equation}
    L_p = \lambda_1 S(I_L,I_L^{'}) + \lambda_2 |I_L - I_{L}^{'}| + \lambda_3 |\nabla I_L - \nabla I_{L}^{'}|
\end{equation}
where we set $\lambda_1=0.85$, $\lambda_2=0.15$, $\lambda_3=0.15$. These values were selected through experimentation.
\item
Regularization loss ($L_r$): Regularization loss is used to smooth local disparity with information directly from input images, and we only use it in estimating initial disparity. We assume disparity in the local region tends to be smooth, so we add a regularization loss to suppress high frequency noise introduced by the photometric loss term. This regularization loss is the sum of the weighted second derivative of the disparity map, and the weight is the exponential of the second derivative of the input image. The higher the second derivative of the input image, the higher the probability of a change in disparity. For the left side, regularization loss is defined as follows:
\begin{equation}
    L_r = \frac{1}{N}\sum |\nabla_{x}^{2}D_L|e^{-|\nabla_{x}^{2}I_L|} + |\nabla_{y}^{2}D_L|e^{-|\nabla_{y}^{2}I_L|}
\end{equation}
where $N$ is number of pixels, $\nabla_{x}^{2}$ and $\nabla_{y}^{2}$ are second derivatives along the X and Y axes.

\item
Consistency loss ($L_c$): We can also synthesize a left image from the reconstructed right image $I_{L}^{''} = F(I_{R}^{'}, D_L)$ and a right image from the reconstructed left image $I_{R}^{''} = F(I_{L}^{'}, D_R)$. Consistency loss is defined as follows:
\begin{equation}
    L_c = |I_L - I_{L}^{''}| + |I_R - I_{R}^{''}|
\end{equation}
This consistency forces the left and right branches to be consistent with one another~\cite{zhong2017self}. 

\item
Smoothness loss ($L_s$): For difficult regions, we argue that the network should be able to infer the disparity from its neighbors within a segment. Assuming initial disparity is reasonable, we use a left-right consistency check to find these regions. So we only include this loss in the refinement. We warp the right disparity $D_R$ using the left disparity $D_L$, and we form a reconstructed image $D_L^{'} = F(D_R, D_L)$. Then we threshold the absolute difference between $D_L$ and $D_L^{'}$: 
\begin{equation}
Diff = 
\begin{cases}
    ||D_L - D_L^{'}||,      & ||D_L - D_L^{'}|| <= t\\
    t,                      & ||D_L - D_L^{'}|| > t
\end{cases}
\end{equation}

where t is the threshold and is set to 3 during the experiment. Too large of a threshold will result in a trivial solution. In addition, the disparity should be smooth inside a segment. These segments are learned from the semantic segmentation task. Shallower layers are used here instead of the final semantic segmentation layer, biasing to smaller segments being learned. We apply a cost to enforce smoothness within a segment. For the left side,
\begin{equation}
\begin{split}
    L_s = \frac{1}{N}\sum |\nabla_{x}^{2}D_L|(e^{-|\nabla_{x}^{2}f_L|} + e^{(Diff-t)})\\
        + |\nabla_{y}^{2}D_L|(e^{-|\nabla_{y}^{2}f_L|} + e^{(Diff-t)})  
\end{split}
\end{equation}
where $f_L$ is feature vectors from the left view. This loss is only applied during refinement because it is conditioned on relatively good initial disparity. \textcolor{blue}{}

\item 
Segmentation loss ($L_{seg}$): Conventional softmax cross entropy loss is used to measure the difference between the logits map and the ground truth segment labels. For the stereo image dataset, only images from one side will be labeled. For example, KITTI and Cityscapes only have segment labels for the left images. However, the left disparity map will relate the left and right images. So we can use the left disparity map to warp the right output segments to the left, and then we can use the left ground truth label for supervision.
\end{itemize}

\subsection{Post Processing}
Simple post processing can be used to improve the final results. Although the loss of smoothness can reduce the effects of occlusion, our model is still prone to error in those regions. Our post processing consists of two steps: left-right consistency check and interpolation. 

After calculating both left and right disparity, we perform a left-right consistency check. For left view images, a pixel will fail the check if the difference between disparity values from the left view and the corresponding pixel from the right view is greater than a certain threshold. We set this threshold to 1, and we end up with a boolean mask. We also apply a median filter to this mask because it contains a fair degree of noise. Then, in these failure regions, we assign them disparity values from the background. As proposed by \citet{zbontar2016stereo}, we interpolate by moving left until finding a position with a valid disparity and use this as its value. No further global optimization is applied.

\section{EXPERIMENTAL EVALUATION}
In this section, we explain our implementation details and present qualitative and quantitative results.

\subsection{Datasets}
\textbf{KITTI}: KITTI 2012 and 2015 are two benchmark real-world driving datasets. They provide ground truth disparity computed from a calibrated high-resolution 3D LIDAR. There are approximately 200 rectified stereo images with ground truth disparity for evaluation in both KITTI 2012 and 2015. We primarily focus on the KITTI 2015 benchmark. Compared to KITTI 2012, challenging regions (e.g. car windshields) from KITTI 2015 are more correctly represented in the ground truth because it uses CAD models to produce disparity values for evaluation. Additionally, only KITTI 2015 contains ground truth for semantic segmentation. For evaluation, pixels are divided into two overlapping categories: strictly non-occluded regions (NOC) and all pixel regions (ALL). The KITTI 2015 benchmark considers a pixel to be "correct" if the disparity error is less than 3 pixels and within 5\% disparity error. 

\textbf{Cityscapes}:  Cityscapes is a dataset for semantic urban scene understanding. It contains 5,000 stereo color images collected from 50 cities, with high quality pixel-level ground truth semantic labels for the left view of each pair. These images are split into sets, with 2,975 for training, 500 for validation and 1,525 for testing. There are no ground truth disparity maps in the Cityscapes dataset, but disparity maps are provided using the SGM~\cite{hirschmuller2008stereo} algorithm.

\subsection{Implementation Details}
In the experiments, we implement our architecture in TensorFlow. All experiments are run on a single NVIDIA Titan-X GPU. Original stereo images are normalized to values ranging from -1 to 1. Due to GPU memory limitation, we have a maximum batch size of 1, the maximum disparity is set to 192 and images are randomly cropped down to 256x512 patches before feeding into network. During optimization, we use the Adam optimizer~\cite{kinga2015method} with $\beta_1=0.9$, $\beta_2=0.999$ and $\epsilon=1e{-8}$. The learning rate is set to $2e{-4}$ for pre-training on Cityscapes and $1e{-4}$ for fine-tuning on KITTI, and it is halved every $20,000$ iterations. Pre-training on Cityscapes is done for $100,000$ iterations. We then fine-tune the model on KITTI for an additional $50,000$ iterations. The finetuning process takes approximately 1 day. No data augmentation is performed in the experiments.

\begin{figure}[t!]
\vspace{5pt}
\centering
\subfloat[Sample results on test set in KITTI 2015. From top: left stereo input image, disparity prediction, error map.]{%
  \includegraphics[clip,width=0.9\linewidth,height=4cm]{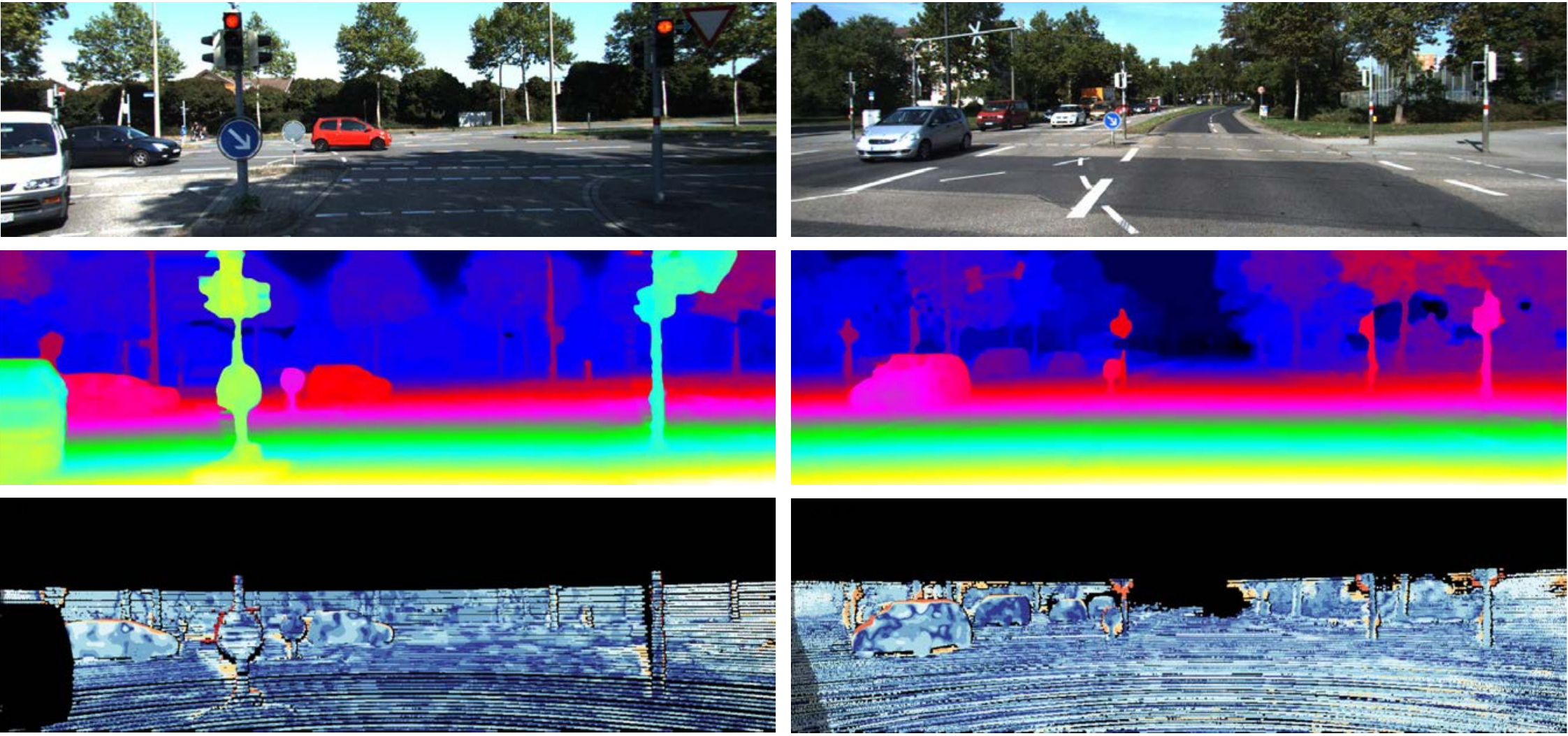}}%
  \vspace{-0.1pt}
\subfloat[Sample results on Cityscapes. From top: left stereo input image, disparity prediction from SGM, disparity prediction from ours. ]{%
  \includegraphics[clip,width=0.9\linewidth,height=4cm]{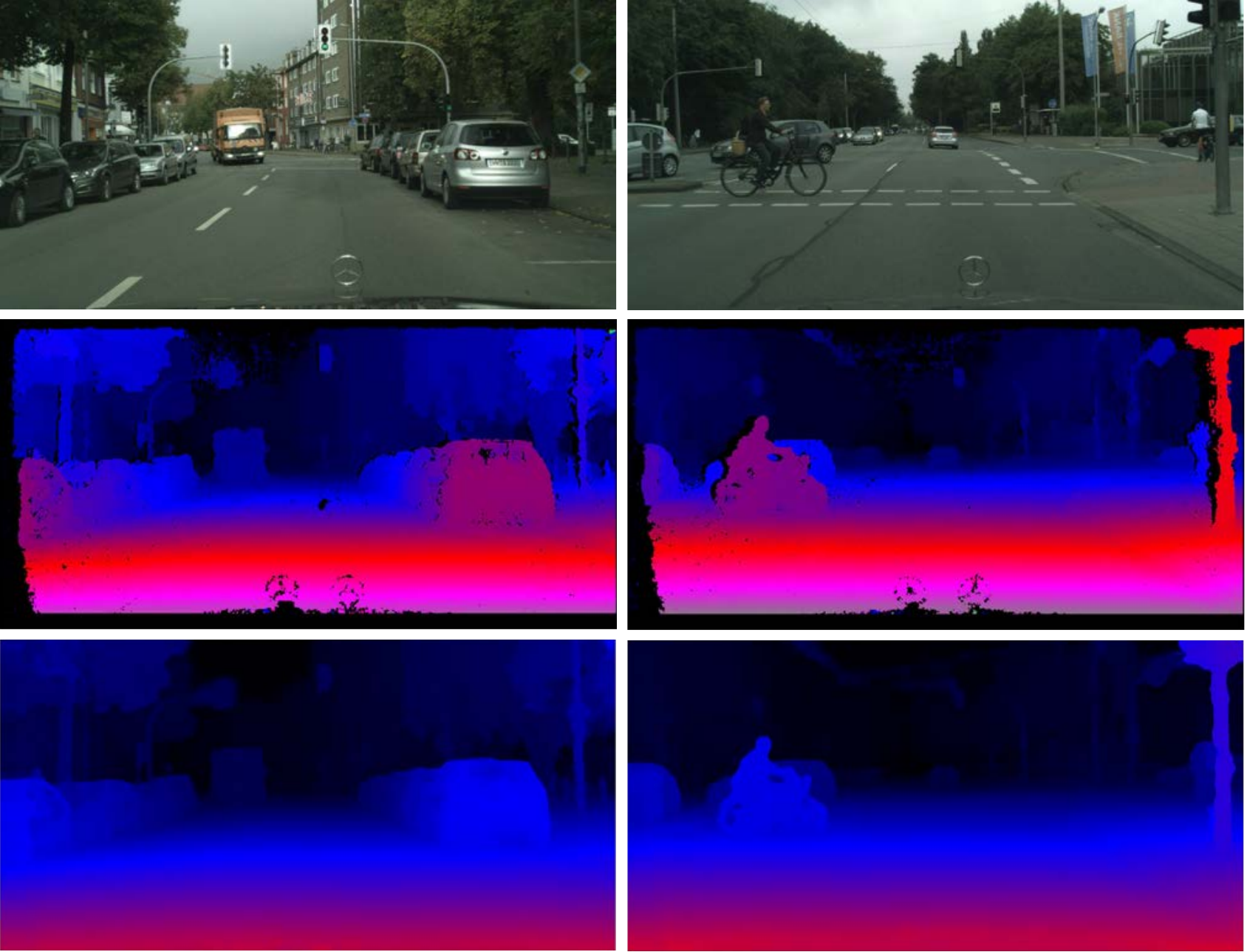}}%
\caption{Qualitative results on KITTI 2015 and Cityscapes datasets. (a)Sample results on KITTI 2015 test set. The last column shows the error maps which are captured from KITTI benchmark. Error regions are displayed in orange color. (b)Sample results on Cityscapes. Compared with predication from SGM, our model generates more smooth and complete disparity map.}
\label{disparity_results}
\vspace{-10pt}
\end{figure}

\subsection{Evaluation}
Here, we report the results of our model on the KITTI and Cityscapes datasets and compare our approach to other state-of-the-art methods. 
\subsubsection{KITTI Benchmark}
We report results on 40 validation images split from 200 training stereo images from KITTI 2015 to evaluate our model. We compare our model with other unsupervised learning methods in Table~\ref{table_comparison}. Note that our model outperforms other unsupervised methods by a notable margin. In the table, 'CS' refers to training model on Cityscapes dataset, 'K' refers to training model on KITTI and 'PP' refers to refining disparity with post processing. With pre-training on the Cityscapes dataset and simple post processing, results of our model are further improved. In addition, Table \ref{table_kitti_2015} compares our method to other supervised approaches on the KITTI 2015 leaderboard. Although there is a gap between performance of current state-of-the-art supervised methods, our model achieves comparable results and even beats DispNet, a supervised method, on background regions. Sample results are shown in Fig. \ref{disparity_results} (a). 

We note that our proposed method has relatively large error in the foreground region. We argue that it is because of significantly larger and more common occlusion and reflection in foreground regions, such as surfaces of vehicles. There exists no correspondence on these regions in the input stereo images. However, unlike other supervised methods that have access to ground truth disparity, our proposed method highly relies on these  correspondence to form photometric loss and uses it as supervision. So it  is reasonable that our method performs poorly on foreground regions. Although semantic segments and post processing have been used to greatly reduce such errors, our method cannot reach the accuracy of those supervised methods. 
\subsubsection{Cityscapes}
We only show qualitative results from the Cityscapes dataset because it does not provide ground truth disparity maps. The results are shown in Fig. \ref{disparity_results}. Note that compared with the SGM approach, our model is able to generate much more complete and visually accurate disparity maps.

\begin{table}[t!]
    \vspace{3pt}
    \begin{center}
    \caption{Comparison with other unsupervised models on disparity estimation. Results are reported on the KITTI 2015 stereo validation set manually splitted from 200 training images. 'CS' refers to training model on Cityscapes dataset; 'K' refers to training model on KITTI; 'PP' refers to refining disparity with post processing. With pretraining on Cityscapes, fine-tuning on KITTI and post processing, our model outperforms other unsupervised methods by a large margin.}
    \begin{tabular}{|c||c|c|}
        \hline
        Model & NOC pixels & All pixels\\
        \hline
        USCNN~\cite{ahmadi2016unsupervised} & 11.17 & 16.55\\
        \hline
        Zhou et al.~\cite{zhou2017unsupervised} & 8.61 & 9.91 \\
        \hline
        Godard et al.~\cite{godard2017unsupervised} & - & 9.19 \\
        \hline
        SegStereo~\cite{yang2018segstereo} & 7.70 & 8.79 \\
        \hline
        Luo et al.~\cite{luo2018unsupervised} & 6.31 & 6.63 \\
        \hline
        \hline
        Ours(CS) & 6.55 & 7.24 \\
        \hline
        Ours(K) & 5.93 & 6.32  \\
        \hline
        Ours(CS \& K) & 5.84 & 6.29  \\
        \hline
        Ours(K \& pp) & 5.29 & 5.69 \\
        \hline
        Ours(CS \& K \& pp) & \textbf{5.20} & \textbf{5.67} \\
        \hline
    \end{tabular}
    \label{table_comparison}
    \end{center}
    \vspace{-10pt}
\end{table}

\begin{table*}[t!]
    \vspace{8pt}
    \begin{center}
    \caption{Comparison with other supervised methods on disparity estimation. Results are reported on KITTI 2015 test set. Numbers indicate the percentage of pixels which have greater than three pixels or 5\% disparity error. 'D1-bg', 'D1-fg' and 'D1-All' refer to background pixels which contain static elements, dynamic object pixels and all pixels respectively. Although there is a gap between performance of supervised methods and ours, our model shows decent results and even beats DispNet on the D1-bg region.}
    \begin{tabular}{c|ccc|ccc|c}
          &       & NOC   &        &       & All   &        &    \\
    Model & D1-bg & D1-fg & D1-All & D1-bg & D1-fg & D1-All & Runtime\\
    \hline\hline
    DispNet~\cite{mayer2016large} & 4.11 & \textbf{3.72} & 4.05 & 4.32 & \textbf{4.41} & 4.34 & \textbf{0.06}\\ 
    Content-CNN~\cite{luo2016efficient} & 3.32 & 7.44 & 4.00 & 3.73 & 8.58 & 4.54 & 1\\ 
    MC-CNN~\cite{zbontar2016stereo} & 2.48 & 7.64 & 3.33 & 2.89 & 8.88 & 3.89 & 67 \\ 
    GC-Net~\cite{kendall2017end} & 2.02 & 5.58 & 2.61 & 2.21 & 6.16 & 2.87 & 0.9\\
    PSMNet~\cite{chang2018pyramid} & \textbf{1.71} & 4.31 & \textbf{2.14} & \textbf{1.86} & 4.62 & \textbf{2.32} & 0.41\\
    \hline
    Ours   & 3.86 & 15.89 & 5.84 & 4.20 & 16.97 & 6.33 & 0.9\\
    \hline
  \end{tabular}
  \label{table_kitti_2015}
  \end{center}
  \vspace{-10pt}
\end{table*}

\subsection{Ablation Study on Loss Components}
We perform ablation experiments to evaluate the different components of our developed loss function. Results of the ablation study are shown in Table~\ref{table_study_loss}. Models are trained and evaluated on the KITTI 2015 without pretraining or any post processing. The results of our model are improved due to the two-stage refinement, designed smoothness loss and incorporation of semantic segmentation supervision. Specifically, the error rate is reduced from 7.04 to 6.53 with designed smoothness loss and is further reduced from 6.53 to 5.93 with segment supervision. Fig. \ref{failure_case} shows a qualitative result. With semantic segmentation supervision, it corrects the wrongly estimated disparity on the center of the road, which is a region with high reflection.

\subsection{Performance Analysis}
In Table \ref{Performance_analysis}, we present details error rates on regions of each semantic segmentation class before and after adding smoothness loss and fusing segment embedding. We wish to delve into how segment embedding learned from semantic segmentation benefits disparity estimation. In the table, 'smo' refers to smoothness loss; 'seg' refers to segmentation loss; the first row shows the name of semantic classes; the second row shows error rates of the model with all components of losses except smooth loss and segmentation loss; the third row shows error rates of the model with all losses except segmentation loss; the fourth row shows error rates of the model with all losses; the final row shows percentage of error rate reduction after adding smooth loss and segmentation loss. As shown in the table, the smoothness loss helps improve disparity estimation for large semantic classes but not for small semantic classes. For example, error rates on regions of large semantic classes like roads, cars and buses decrease substantially, but error rates on regions of small semantic classes, such as poles, traffic lights and traffic signs, actually increase after imposing smoothness loss. This is because, without guidance of semantic segmentation, smoothness loss tends to blindly force local disparity smooth and disparities for small objects are smoothed to their neighbors which results in more error. 

However, with supervision of the semantic segmentation task, the model is able to learn semantic features. In this case, disparity smoothness loss will force the disparity to be smooth within segments with the same semantic meanings rather than blindly with neighboring segments. Thus, disparities for small objects will remain coherent. It is shown in the table that error rates on regions of poles, traffic lights, traffic signs and other small semantic classes decrease to the lowest level after supervision of the semantic segmentation task.

The focus of this work is on improving state-of-the-art for unsupervised disparity estimation guided by semantic segmentation. We also evaluate our method on semantic segmentation performance. Our baseline IoU is 47.6\%. After disparity refinement, segmentation performance decreases slightly to 46.9\% when evaluating on 40 validation images from KITTI 2015. This suggests that the disparity loss forces features to be different even within a semantic class.

\begin{table*}[t!]
    \centering
    \caption{Error rates of disparity estimation on regions of each semantic class. In the table, 'smo' refers to smoothness loss; 'seg' refers to segmentation loss; the first row shows the name of semantic classes; the second row shows error rates of the model with all components of losses except smoothness loss and segmentation loss; the third row shows error rates of the model with all losses except segmentation loss; the fourth row shows error rates of the model with all losses; the final row shows percentage of error rate reduction after adding smoothness loss and segmentation loss. Smoothness loss improves performance on relatively large semantic classes but not on small semantic classes. With supervision of the semantic segmentation task, error rates on regions of small semantic classes decrease substantially.}
    \small
    \setlength\tabcolsep{2pt}
    \begin{tabular}{c|ccccccccccccccccccc}
    \hline
    Method & road & pole & car & tsign & bus & swalk & train & wall & build. & tlight & veg. & fence & truck & person & bike & terrain & rider & mbike \\
    \hline \hline   
    Model & 2.65 & 11.26 & 12.94 & 6.33 & 2.31 & 5.53 & 1.25 & 2.45 & 14.82 & 2.34 & 10.60 & 11.34 & 6.23 & 1.52 & 2.51 & 6.06 & 1.06 & 0.34 \\ 

    Model(smo) & 1.51 & 13.13 & 10.53 & 6.35 & 1.85 & 4.36 & 1.15 & 2.28 & 13.53 & 2.63 & 9.57 & 11.15 & 6.26 & 1.50 & 2.40 & 5.27 & 1.24 & \textbf{0.33} \\

    Model(smo\&seg) & \textbf{1.35} & \textbf{7.62} & \textbf{8.83} & \textbf{4.67} & \textbf{1.77} & \textbf{4.25} & \textbf{0.99} & \textbf{2.09} & \textbf{13.05} & \textbf{2.06} & \textbf{9.36} & \textbf{10.13} & \textbf{5.76} & \textbf{1.43} & \textbf{2.36} & \textbf{5.81} & \textbf{1.04} & 0.36 \\
    \hline
    Improvement \% & 48.88 & 32.37 & 31.72 & 26.20 & 23.29 & 23.09 & 20.37 & 14.40 & 11.94 & 11.75 & 11.64 & 10.73 & 7.50 & 6.07 & 5.95 & 4.21 & 2.50 & -6.38 \\
    \hline
    \end{tabular}
    \label{Performance_analysis}
\end{table*}

\begin{table*}[t!]
  \centering
  \caption{Ablation study on loss components. Results of models with different losses are reported on KITTI 2015 training set without pretraining or post processing. There are two stages in our model. The superscript 'init' refers to losses in the initial stage and the superscript 'ref' refers to losses in the refinement stage. The subscript 'p' refers to photometric loss, 'r' refers to regularization loss, 'c' refers to consistency loss, 's' refers to smoothness loss and 'seg' refers to supervised semantic segmentation loss. Results shown here justify our two-stage architecture and designed components of total loss.}
  \begin{tabular}{ccccccc|c|c}
    \hline
    $L_p^{init}$ & $L_c^{init}$ & $L_r^{init}$ & $L_p^{ref}$ & $L_c^{ref}$ & $L_s^{ref}$ & $L_{seg}$ & NOC pixels & All pixels\\
    \hline\hline
    $\surd$ & $\surd$ & $\surd$ &   &   &   &   & 7.18 & 8.75\\ 
    $\surd$ & $\surd$ & $\surd$ & $\surd$ & $\surd$ &   &   & 7.04  & 8.60\\ 
    $\surd$ & $\surd$ & $\surd$ &   &   &   & $\surd$  & 6.70  & 8.14\\ 
    $\surd$ & $\surd$ & $\surd$ & $\surd$ & $\surd$ & $\surd$ &     & 6.53 & 6.94\\
            &         &   & $\surd$ & $\surd$ & $\surd$ & $\surd$ & 5.99  &  6.42\\
    $\surd$ & $\surd$ & $\surd$ & $\surd$ & $\surd$ & $\surd$ & $\surd$ & \textbf{5.93} & \textbf{6.32}\\
    \hline
  \end{tabular}
  \label{table_study_loss}
  \vspace{-5pt}
\end{table*}

\subsection{Qualitative Results: 3D Models}
\begin{figure}[t!]
\vspace{4pt}
\centering
\subfloat[Sample 3D semantic results on KITTI 2015. From top: left stereo input images, 3D cloud points, semantic segmentation on 3D point clouds.]{%
  \includegraphics[clip,width=\linewidth,height=4cm]{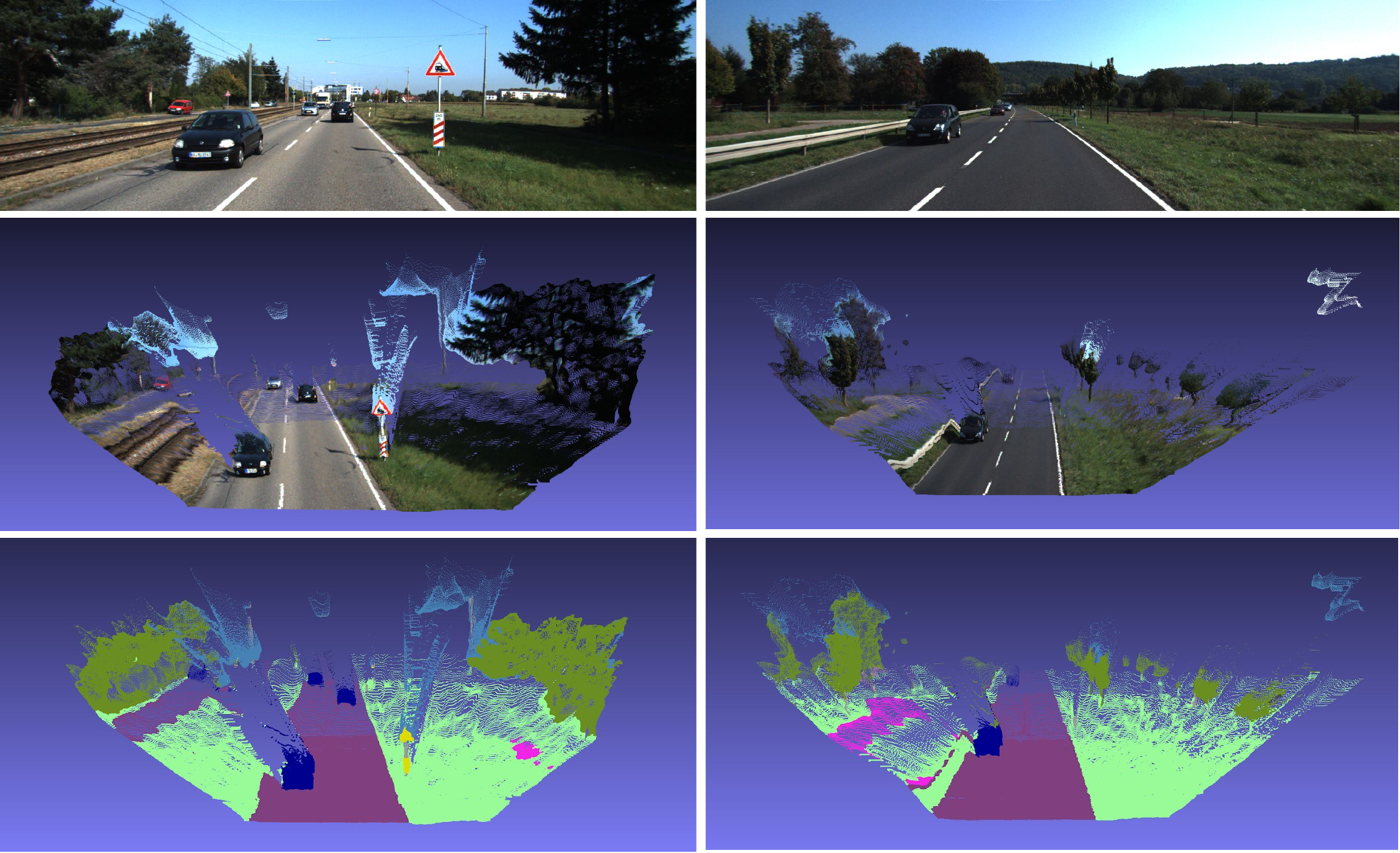}}%
  \vspace{-0.5pt}
\subfloat[Sample 3D semantic results on Cityscapes. From top: left stereo input images, 3D cloud points, semantic segmentation on 3D point clouds.]{%
  \includegraphics[clip,width=\linewidth,height=4cm]{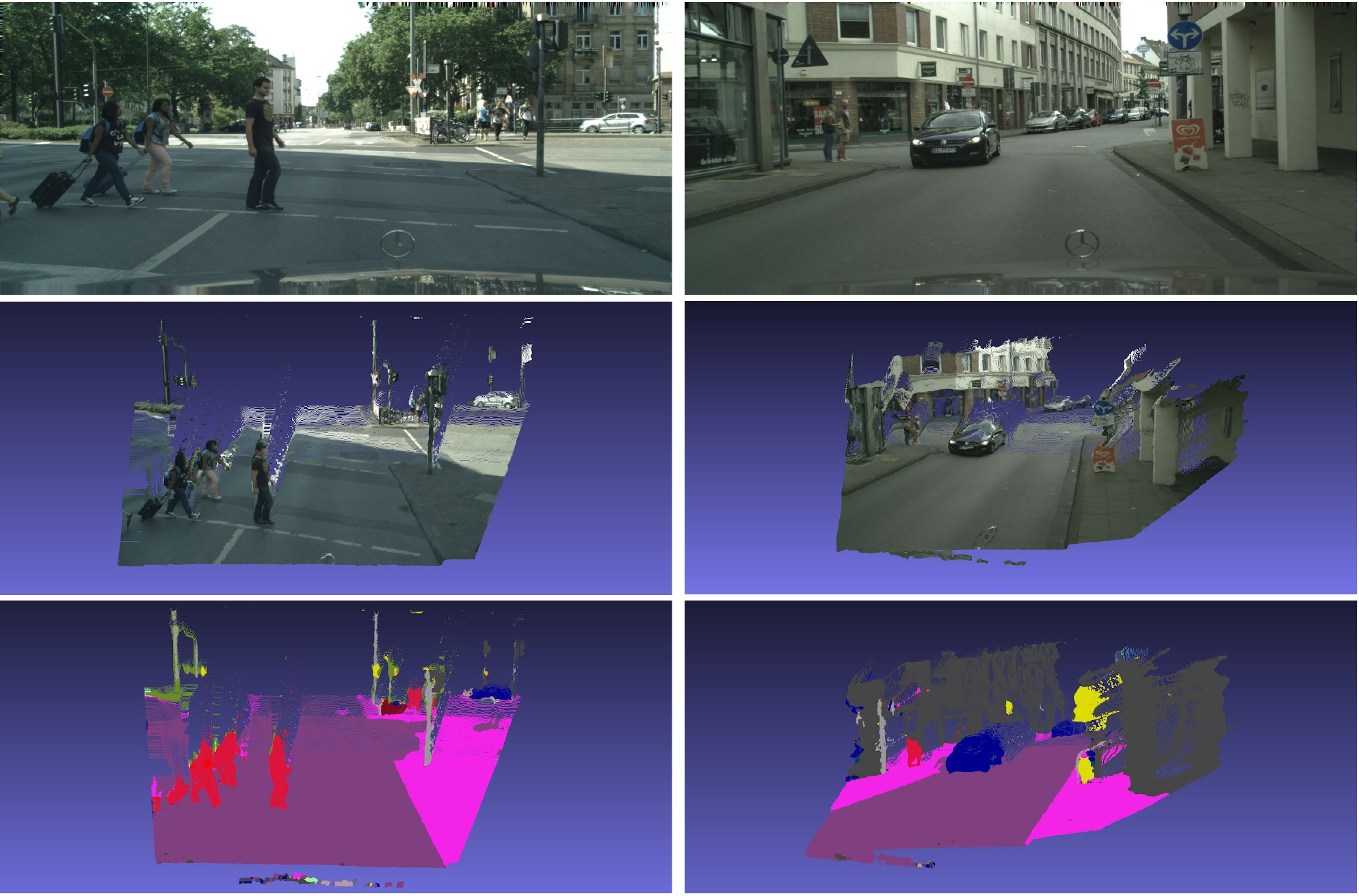}%
}
\caption{Sample qualitative 3D semantic results on KITTI 2015 and Cityscapes datasets. (a) 3D Reconstruction Results on KITTI 2015. (b)3D Reconstruction Results on Cityscapes. The last rows in both (a) and (b) show the 3D semantic results. Different color refers to different semantic class.}
\label{3D_reconstruction}
\vspace{-15pt}
\end{figure}

We triangulate the disparity maps with the camera extrinsics into 3D point clouds with semantic labels as shown in Fig.~\ref{3D_reconstruction}. We only consider pixels where disparities are above 5. Note that simultaneously calculating both disparity and semantic class enables us to efficiently produce semantic 3D models, which can be used more directly for driving tasks than other independent outputs.





\section{CONCLUSIONS}
We propose a model in which segment embedding learned from semantic segmentation is fused into the process for disparity estimation. This segment embedding is helpful for estimating disparity in ill-posed regions. We demonstrate the efficacy of our method on both KITTI and Cityscapes datasets. Our unsupervised method achieves comparable results to supervised methods on KITTI and even outperforms some of them in background regions. Outputting disparities and semantic segments simultaneously enables us to efficiently produce semantic 3D models. For future work, we are going to exploit instance segment labels, as instance segments have potential to provide further cues for object boundaries and finer details.


\section*{ACKNOWLEDGMENT}
This work was supported by a grant from Ford Motor Company via the Ford-UM Alliance under award N022884.

\printbibliography

\end{document}